\documentclass[9pt,twocolumn,twoside]{osajnl}

\usepackage{times}

\usepackage{amsmath}
\usepackage{amssymb}

\usepackage{bm}
\usepackage{multirow}
\usepackage{mathtools}
\usepackage{units}
\usepackage{subfloat}
\usepackage{float}
\usepackage{xcolor}
\usepackage{url}
\usepackage{bbm}
\usepackage{algorithm}
\usepackage{algpseudocode}
\usepackage{pbox}
\usepackage{placeins}
\usepackage{tabularx} %
\usepackage{rotating}
\usepackage{subcaption}
\usepackage{appendix}
\usepackage{stfloats}

\usepackage{booktabs}
\usepackage{multirow}
\usepackage{tabularx}
\newcolumntype{C}{>{\centering\arraybackslash}X}
\newcolumntype{x}[1]{>{\centering\let\newline\\\arraybackslash\hspace{0pt}}p{#1}}

\usepackage[TS1,T1]{fontenc}

\usepackage{graphicx}

\makeatletter
\renewcommand{\fnum@figure}{\textbf{Figure \thefigure}}
\renewcommand{\fnum@table}{\textbf{Table \thetable}}
\makeatother

\usepackage{bm}

\newcommand{\wfr}[0]{\ensuremath{\mathcal{W}}} %
\newcommand{\bfr}[0]{\ensuremath{\mathcal{B}}} %

\usepackage{tikz,pgfplots}
\usetikzlibrary{matrix}
\pgfplotsset{compat=1.15}
\usepackage{etex} 
\usetikzlibrary{arrows,automata}
\usetikzlibrary{positioning}
\usepgfplotslibrary{groupplots}
\usepackage{tikz-3dplot}
\tikzset{
	state/.style={
		rectangle,
		rounded corners,
		draw=black, very thick,
		minimum height=2em,
		inner sep=2pt,
		text centered,
	},
}
\tikzset{
	info/.style={
		rectangle,
		draw=black, thin,
		minimum height=2em,
		inner sep=2pt,
		text centered,
	},
}
\usetikzlibrary{positioning}
\usetikzlibrary{calc}
\usetikzlibrary{arrows,shapes,backgrounds}

\definecolor{ours}{rgb}{0.4660, 0.6740, 0.1880}
\definecolor{fastplanner}{rgb}{0.85,0.325,0.098}
\definecolor{reactive}{rgb}{0,0.447,0.741}
\definecolor{blind}{rgb}{1.0, 0.0, 0.0}
\definecolor{m_yellow}{rgb}{0.929,0.694,0.125}
\definecolor{m_purple}{rgb}{0.4940, 0.1840, 0.5560}

\journal{ol} 

\setboolean{shortarticle}{false}

\title{Superhuman Safe and Agile Racing through \\Multi-Agent Reinforcement Learning}

\author[*1]{Ismail Geles}
\author[1]{Leonard Bauersfeld}
\author[2]{Markus Wulfmeier\thanks{Research primarily conducted while at Google DeepMind; currently at Nomagic.}}
\author[1]{Davide Scaramuzza}

\affil[1]{Robotics and Perception Group, University of Zurich, Zurich, Switzerland}
\affil[2]{Google DeepMind, London, United Kingdom}
\affil[*]{Corresponding author: geles@ifi.uzh.ch}

\begin{abstract}
Autonomous systems have achieved superhuman performance in isolation or simulation, yet they remain brittle in shared, dynamic real-world spaces. This failure stems from the dominant single-agent paradigm for physical applications, where other actors are ignored or treated as environmental noise, preventing effective coordination.
Here we show that multi-agent reinforcement learning provides the essential safety scaffolding required for real-world interaction. Using high-speed quadrotor racing as a high-stakes testbed, we train agents to navigate complex aerodynamic interactions and strategic maneuvering with a variable number of racers. Through league-based self-play, agents evolve sophisticated anticipatory behaviors, including proactive collision avoidance, overtaking, and handling multi-agent physical interactions, including aerodynamic downwash.
Our agents outperform a champion-level human pilot in multi-player races at speeds exceeding 22\,m/s, while simultaneously reducing collision rates by 50\% compared to state-of-the-art single-agent baselines. 
Crucially, training with diverse artificial agents enables zero-shot generalization to safer human interaction.
These results suggest that the path to robust robotic co-existence lies not in isolated safety constraints, but in the rigorous demands of multi-agent interaction. 
Multimedia materials are available at:  \url{https://rpg.ifi.uzh.ch/marl}
\end{abstract}

\begin{document}

\maketitle

\section*{Introduction} \label{sec:introduction}
\noindent
Autonomous robots are increasingly operating in the physical world. Learned policies now enable walking robots to traverse challenging terrain~\cite{miki2022learning, lee2020learning}, mobile manipulators to clean up~\cite{wu2023tidybot} or cook~\cite{fu2024mobile}, and aerial vehicles to navigate complex environments at high speed~\cite{loquercio2021learning, kaufmann2023champion, song2023reaching, bahnam2026monorace, geles2024demonstrating}. 
Yet, a striking asymmetry persists: while single-agent systems have matured from laboratory demonstrations to practical deployment, agile multi-agent coordination remains largely confined to simulation~\cite{gronauer2022multiagent, cusumanotowner2025robust}, even in highly realistic and competitive domains such as autonomous car racing~\cite{wurman2022gtsophy}. 

In the physical world, head-to-head racing against a human opponent does not resolve this challenge either: it remains fundamentally a duel, a bounded interaction in which strategy largely reduces to lap time optimization. Once the autonomous agent establishes a lead, the risk of collision with its single opponent effectively vanishes, as empirically observed in prior head-to-head racing against champion-level humans, where the autonomous agent led from start to finish in every race it completed~\cite{kaufmann2023champion}. When additional competitors enter the same airspace, this simplification breaks down. Strategies optimized for solo or head-to-head racing suffer sharp performance degradation as the number of agents grows, with increasing collisions and deteriorating race-completion rates. Strategies that master a duel, lose in multi-player interaction and deployment. This gap presents a fundamental barrier to scaling autonomous systems, as many critical applications, from warehouse logistics to search and rescue to urban air mobility, inherently require multiple robots operating in shared physical space alongside humans. Addressing this challenge demands agents which model, anticipate, and adapt to the behavior of multiple other agents sharing the same physical space.

Artificial intelligence has encountered this transition from single- to multi-opponent settings before, and each instance has required new methods rather than incremental scaling. In poker, AI first achieved superhuman play in two-player zero-sum settings~\cite{bowling2015solving, brown2018superhuman}, where the Nash equilibrium provides strong worst-case guarantees and remains computationally tractable. The transition to six-player poker proved categorically harder: equilibrium strategies no longer guarantee bounded exploitability, and the combinatorial growth of opponent interactions demands qualitatively different search methods to achieve superhuman performance~\cite{brown2019superhuman}.
Just as game-playing AI had to advance from two-player contests to multiplayer settings, robotic control must advance from optimizing solo trajectories to dynamic co-existence of multiple agents in shared physical environments. 

Reinforcement learning has a long history of superhuman game play~\cite{tesauro1995temporal}, from backgammon and Go~\cite{silver2016mastering, silver2017mastering} to the partial observability and vast strategy spaces of StarCraft~II~\cite{vinyals2019grandmaster}, Dota~2~\cite{berner2019dota}, and multi-player capture-the-flag~\cite{jaderberg2019human}.  A key insight from these systems is that training against a diverse population of opponents, rather than a single fixed adversary, prevents strategic overfitting and produces more generalizable behaviors~\cite{lanctot2017unified, balduzzi2019open, cusumanotowner2025robust}, including interaction with humans \cite{silver2016mastering, vinyals2019grandmaster, berner2019dota}. Beyond game play, reinforcement learning has produced emergent team behaviors in simulated robotics %
settings~\cite{liu2022motor, baker2020emergent, bansal2018emergent}. Consequently, recent work has begun to bridge the gap to physical systems through sim-to-real transfer, demonstrating competitive behaviors in robot soccer~\cite{haarnoja2024learning, tirumala2024learning}, small-scale car racing~\cite{werner2023dynamic}, and two-player drone racing with sparse competitive rewards~\cite{pasumarti2025agileflightemergesmultiagent}. 

However, these systems either operate at low speeds and limited safety challenges, or are restricted to two-player interactions that do not capture the combinatorial complexity of multi-agent coordination.
Translating multi-agent learning to more agile physical platforms compounds the well-studied challenges of non-stationarity, exponential state-space growth, and partial observability~\cite{busoniu2008comprehensive, gronauer2022multiagent} with constraints unique to the physical domain: collisions destroy hardware, %
interactions between agents couple their dynamics, and safety cannot be compromised for performance. Whether multi-agent learning can enable safe coordination at speeds where collisions are catastrophic and reaction margins are measured in milliseconds remains an open problem.

Multi-player quadrotor racing provides a uniquely demanding testbed for this question. The domain demands performance at the physical limits of vehicle dynamics, with speeds exceeding $\unit[100]{km/h}$ and accelerations surpassing $\unit[12]{g}$~\cite{song2023reaching}. Prior autonomous racing systems have reached these limits but optimize purely for lap time regardless of competitive context~\cite{kaufmann2023champion, song2023reaching, bahnam2026monorace}. Existing multi-agent aerial systems have demonstrated coordination capabilities but operate at speeds up to an order of magnitude below competitive racing~\cite{preiss2017crazyswarm, spica2020real, zhou2022swarm}. When multiple vehicles share the track at high speeds, the problem changes significantly. An $N$-agent race creates a densely coupled system in which every trajectory constrains every other. Gates become contested narrow passages where vehicles must negotiate entry. Aerodynamic downwash from neighboring vehicles perturbs flight dynamics in ways that cannot be predicted from single-agent experience. The challenge shifts from pure speed to a negotiation between performance and survival, requiring the capacity to anticipate, yield to, or exploit the behavior of multiple competitors simultaneously.

Here we present a multi-agent reinforcement learning framework that addresses this challenge, achieving superhuman safe multi-player quadrotor racing at speeds reaching $\unit[22]{m/s}$ and accelerations of up to $\unit[7]{g}$. Central to our approach is league-play, inspired by league training~\cite{vinyals2019grandmaster} and fictitious self-play~\cite{heinrich2015fictitious, heinrich2016deep}. Agents train against a diverse, evolving population of competitors drawn from past training checkpoints and from policies obtained through alternative regimes, including single-agent and independent multi-agent learning. A permutation-invariant attention encoder based on Perceiver~\cite{jaegle2021perceiver} allows each agent to process observations invariant to number and order of competitors, while a particle-based downwash model captures the aerodynamic interactions occurring in close-proximity flight. Through competitive self-play, anticipatory behaviors emerge without explicit programming: agents learn to block opponents, yield when overtaking is unsafe, and account for the aerodynamic wake of nearby vehicles, discovering the physics of multi-agent interaction through experience rather than from equations. 
We validate the framework in real-world races with up to four competitors. Compared to single-agent baselines, league-play training reduces collisions by 50\% while preserving competitive lap times, and generalizes beyond the number of competitors seen during training. 
Policies transfer zero-shot to races against human expert pilots, achieving over 90\% race completion rates. These results demonstrate that multi-agent reinforcement learning can extend autonomous aerial systems from isolated operation to safe, high-speed coordination in shared physical environments.

\section*{Results} \label{sec:experiments}

We evaluate our approach in both real-world and simulated multi-agent settings. We use a single policy trained with four agents that generalizes to varying numbers of opponents. In physical experiments, we race with up to 4 competitors, including a champion-level human pilot. The policy maintains both safety and competitive performance across configurations, from solo time trials to dense four-agent races. 
In simulation, large-scale evaluations with diverse opponents and as many as eight agents demonstrate 50\% fewer collisions than single-agent or independently trained policies, highlighting the safety benefits of interaction-aware training. Finally, we examine emergent behaviors such as strategic overtaking and opponent awareness through visualizations of the learned value function.

\subsection*{Champion-level multi-player racing}

Physical deployment of our multi-agent framework is validated through racing experiments spanning time trials, AI-only races, and mixed human-AI competitions against Marvin Schaepper, five-time Swiss national drone racing champion. These experiments assess policies learned entirely in simulation through league-based self-play, evaluating both competitive performance and safe interaction at speeds reaching $\unit[22]{m/s}$ and accelerations of $\unit[7]{g}$.

Experimental configurations range from solo time trials to four-agent races. All platforms share identical hardware: 220-gram, 3-inch quadrotors with a static thrust-to-weight ratio of 6.5, competing on the Split-S racetrack, a 75-meter circuit with seven gates established in prior autonomous racing studies~\cite{kaufmann2023champion, song2023reaching}. The human pilot operates using standard first-person-view equipment consistent with professional racing practice, while autonomous agents execute learned policies using motion-capture state estimation. Each agent observes its own state alongside the relative positions and velocities of all competitors.

\begin{table*}[htb]
\centering
\caption{\textbf{Real-world racing performance across experimental configurations.} Comparison between league-play policy (LP) and expert human pilot M.~Schaepper, five-time Swiss national drone racing champion. Race completion measures the fraction of the three-lap race (21 gates) completed before termination due to collision. Each race yields one trial per pilot.}
\label{tab:race_results}
\setlength{\tabcolsep}{4pt}
\begin{tabularx}{1\linewidth}{llx{1.5cm}x{3cm}CC}
\toprule
\textbf{Configuration} &  \textbf{Pilot} & \textbf{Number of Trials} &    \textbf{Avg. Race Completion [\%]}       & \textbf{Best time (1$^\text{st}$ lap) [s]}  & \textbf{Best time-to-finish [s]} \\
\midrule
\multirow{2}{*}{Time Trial}        & LP (ours)      & 13 & 100.00          & \textbf{5.540} & \textbf{16.065} \\
                                   & M. Schaepper   & 10 & 100.00          & 6.627          & 17.109          \\
\midrule
AI Only (2 agents)                 & LP (ours)      &  6 & 100.00 & 5.651 & 16.308 \\
AI Only (3 agents)                 & LP (ours)      &  9 & 100.00 & 5.624 & 16.167 \\
AI Only (4 agents)                 & LP (ours)      & 12 &  90.87 & 6.079 & 16.811 \\
\midrule
\multirow{2}{*}{AI vs Human (1v1)} & LP (ours)      &  5 & \textbf{100.00} & \textbf{5.763} & \textbf{16.280} \\
                                   & M. Schaepper   &  5 &  53.33          & 6.056          & 17.087          \\
\multirow{2}{*}{AI vs Human (3v1)} & LP (ours)      & 12 & \textbf{91.67}  & \textbf{5.998} & \textbf{16.639} \\
                                   & M.~Schaepper   &  4 &  58.33          & 6.554          & 25.776          \\
\bottomrule
\end{tabularx}%
\end{table*}

Table~\ref{tab:race_results} summarizes performance across all experimental conditions. Average race completion serves as the primary metric for safety and robustness, measuring the fraction of the race completed before termination due to collision. A value of 100\% corresponds to completing three laps, requiring successful passage through 21 gates. Lower values indicate earlier termination, reflecting the agent's ability to sustain safe, collision-free flight under competitive pressure.

\begin{figure*}[htbp]
    \centering
    \includegraphics[width=\textwidth]{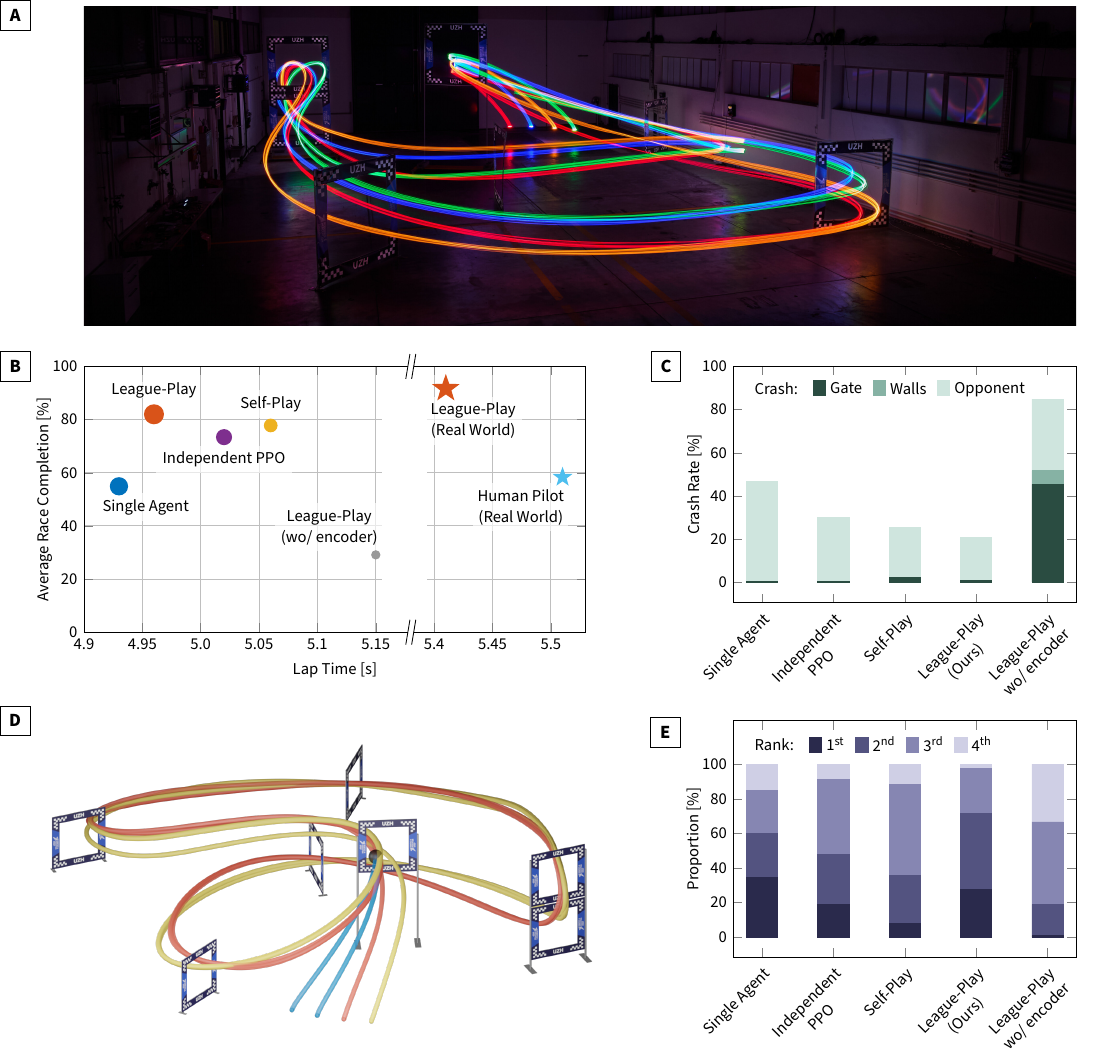}
    \caption{\textbf{A}, Long-exposure photograph of real-world deployment with four agents competing simultaneously. \textbf{B}, Large-scale evaluation of over 64,000 simulated four-player races comparing average lap time against race completion rate. Real-world data points indicate median lap times from all four-player races for our policy and the expert human pilot. \textbf{C}, Crash rates from the large-scale evaluation in B, classified by collision type: gate, wall, or opponent. \textbf{D}, Sample race from the evaluation in B illustrating typical failure modes: two single-agent policies (blue) collide before the first gate, while league-play (orange) and self-play (yellow) continue racing. \textbf{E}, Distribution of finishing positions per method across all races from the large-scale evaluation in B.}
    \label{fig:training_strategies}
\end{figure*}

In time trial configuration, our league-play trained policy completed all 13 trials, achieving the fastest time for the first lap of $\unit[5.540]{s}$ compared to the human champion's $\unit[6.627]{s}$, and the fastest race completion of \unit[16.065]{s} versus $\unit[17.109]{s}$. Notably, both the human pilot and autonomous policy achieved 100\% race completion in isolation, with zero collisions across all trials.
Moreover, in the AI-only races shown in Figure~\ref{fig:training_strategies}A, our league-play policy maintained over 90\% race completion with up to 4 players, highlighting its inherent ability to avoid collisions while competing.

The benefits of interaction-aware training become apparent under multi-agent competition. In one-versus-one races, our policy maintained 100\% race completion across five trials, while the human pilot averaged only 53.33\%. This performance gap suggests that competitive pressure induces riskier behavior in human pilots, a pattern absent in our learned policies. In the more demanding configuration of three autonomous agents racing alongside the human pilot, the learned policies sustained 91.67\% completion compared to 58.33\% for the human, despite the increased density of nearby competitors.

Across all configurations, the league-play policy finished ahead of the human pilot in nearly every race. The sole exception occurred in one of the three-versus-one trials, where a collision between two autonomous agents allowed the human pilot to finish second. In one-versus-one racing, a single mid-air contact occurred between an autonomous agent and the human pilot, while no human-AI collisions were recorded in the three-versus-one configuration.
Analysis of race trajectories reveals a consistent pattern: the human pilot, typically trailing the autonomous agents, attempted increasingly aggressive maneuvers to close the gap, often resulting in gate collisions or loss of control. The learned policies, by contrast, maintained consistent safety margins regardless of competitive standing while using the given opportunities to overtake. These results suggest that interaction-aware training produces policies that internalize robust collision avoidance independent of race dynamics, whereas human pilots modulate risk-taking based on competitive pressure. 
This property is essential for the future deployment of autonomous agents in shared environments, where predictable and conservative behavior around human operators is a prerequisite for safe integration.

\begin{figure*}[htbp]
    \centering
    \includegraphics[width=\textwidth]{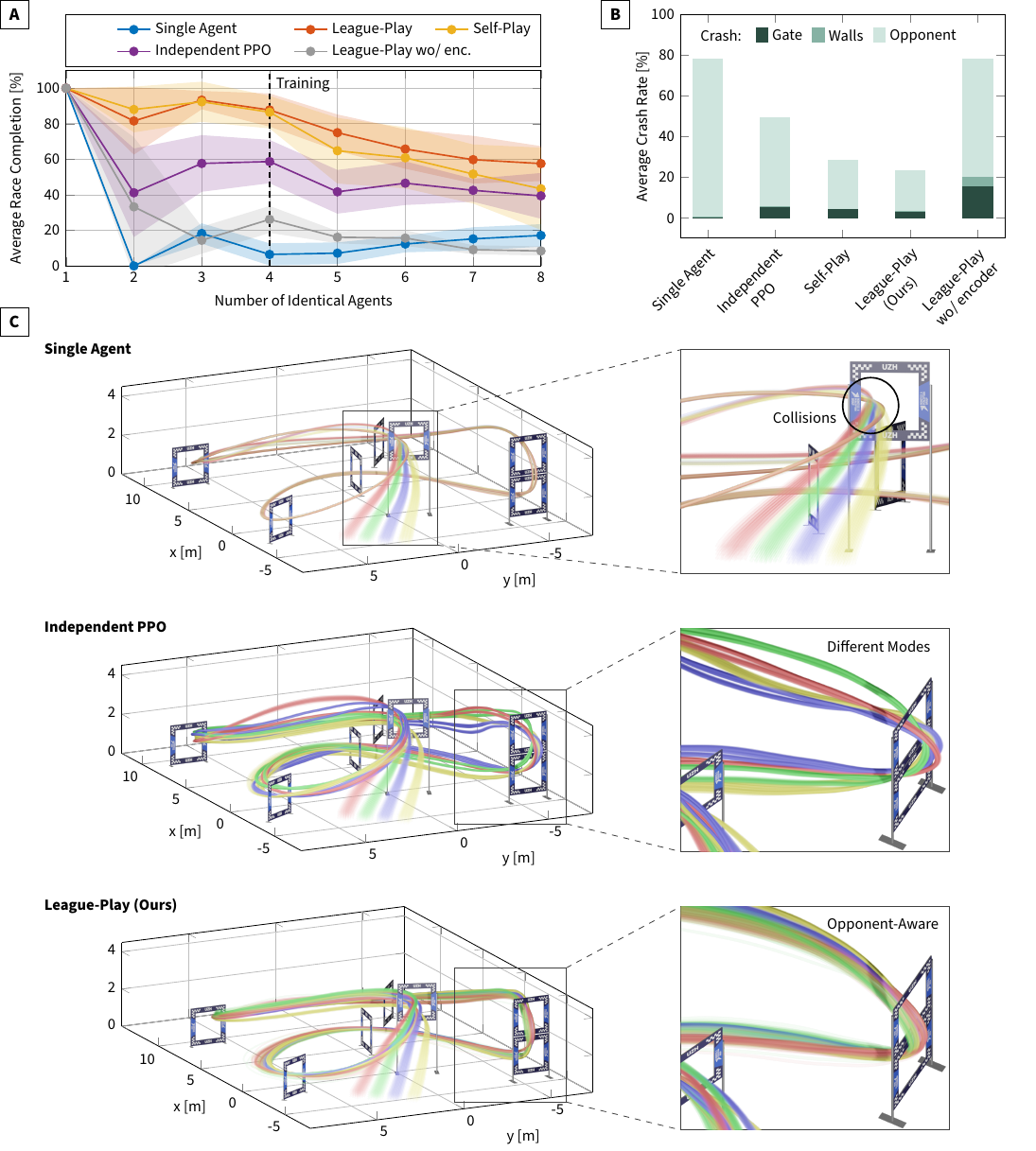}
    \caption{\textbf{A}, Average race completion through self-evaluation with identical policies from solo to 8-player races. Each data point represents four policies per method, each completing 64 races with varied starting positions. Shaded regions indicate the average of standard deviations per policy. \textbf{B}, Crash rates by collision type (gate, wall, opponent) from the races completed in A. \textbf{C}, Sample races illustrating behavioral differences. Single-agent policies collide with 85\% probability at the first gate in four-player races. The independent PPO example shows agents from the same training run achieving collision-free racing through implicit coordination; this behavior does not generalize, dropping to approximately 60\% race completion in self-evaluation. League-play trains against diverse opponents, producing strategies that remain robust across agent configurations.}
    \label{fig:self_eval}
\end{figure*}

\subsection*{Large-scale simulated competitions}

To evaluate generalization and robustness beyond the training distribution, we conducted large-scale simulated competitions across 64,000 four-player races. This evaluation tests our approach against diverse opponent behaviors, including strategies never encountered during training, and enables systematic comparison across training paradigms under controlled conditions.

Five training strategies are compared, all trained with Proximal Policy Optimization (PPO)~\cite{schulman2017proximal} using recurrent actors and critics with LSTM layers~\cite{hochreiter1997lstm} to maintain temporal context: (i) single-agent policies trained without opponent observations, serving as a performance upper bound for lap time but lacking opponent awareness; (ii) independent PPO, where four agents are jointly optimized with opponent observations; (iii) fictitious self-play, where agents train against a history of their own prior checkpoints; (iv) league-play, which augments fictitious self-play with a fixed pool of single-agent and independent PPO opponents; and (v) league-play without the Perceiver-based attention encoder, isolating the contribution of the architecture. All multi-agent methods except the ablation utilize the Perceiver encoder for processing competitor observations. Full training details are provided in the Materials and Methods section.

For fair comparison, we include four different checkpoints from each training strategy to form the evaluation pool. From this pool, we sampled 1,000 opponent configurations, with each configuration executed across 64 races with a grid of varying starting positions such that all competitors begin at equal distances from the first gate. This protocol yields diverse competitive scenarios, enabling direct performance comparison under identical conditions. Results are summarized in Figure~\ref{fig:training_strategies}.
A clear trade-off between speed and safety emerges across training paradigms in Figure~\ref{fig:training_strategies}B. Single-agent policies achieve the fastest average lap time of $\unit[4.93]{s}$ but exhibit the highest crash rates, failing to complete the majority of races. Independent PPO ($\unit[5.02]{s}$) improves collision avoidance relative to single-agent baselines but remains brittle against unfamiliar opponent strategies. Fictitious self-play ($\unit[5.06]{s}$) achieves approximately 80\% race completion, indicating substantially safer behavior at moderate cost to lap time. League-play achieves the highest race completion among all methods while maintaining an average lap time of $\unit[4.96]{s}$, only $\unit[0.03]{s}$ slower than the single-agent baseline. 
This result demonstrates that diverse opponent exposure during training produces policies that generalize safely to novel competitive scenarios with minimal performance cost, validating the core premise that interaction-aware training is essential for safe multi-agent deployment.\\
The crash rate analysis in Figure~\ref{fig:training_strategies}C reveals that opponent collisions dominate across all methods, with gate and wall collisions contributing only a small fraction of total failures. This finding underscores the importance of interaction-aware training: environmental obstacles pose a relatively minor challenge compared to the complexity of avoiding dynamic competitors. This observation aligns with established findings in multi-agent reinforcement learning, where the non-stationarity introduced by interacting agents presents a fundamentally harder learning problem than navigating static environments~\cite{gronauer2022multiagent, busoniu2008comprehensive}.

The ablation without the Perceiver-based attention encoder presents a notable exception: this variant exhibits higher crash rates into gates than into opponents, suggesting that naive concatenation of opponent observations disrupts overall policy performance. Because concatenation assumes fixed-size, ordered inputs, missing or arbitrarily ordered observations degrade behavior substantially. This confirms that the attention-based encoder is essential for safe multi-agent coordination.
Figure~\ref{fig:training_strategies}D illustrates a simulated race in which two single-agent policies collide at the first gate.

The distribution of final race rankings in Figure~\ref{fig:training_strategies}E further illustrates the trade-off between aggressive and safe strategies. Single-agent policies achieve the highest proportion of first-place finishes, optimizing purely for speed without yielding to competitors. However, this apparent advantage is offset by disproportionately high crash rates: single-agent policies, trained purely for time-optimal flight without explicit interaction-awareness, fail to finish far more often than other methods.
League-play, by contrast, rarely ranks last and maintains the most consistent placement across the top positions, demonstrating that interaction-aware training enables competitive performance without the catastrophic failure modes of opponent-agnostic approaches.
In a winner-takes-all setting where only first place matters, the higher win rate of time-optimal (single-agent) policies may outweigh their higher crash rate; under any scoring that rewards consistent finishing, league-play is preferred. The reward formulation affords a degree of control over this balance through the proximity coefficient, see Supplementary Fig.~\ref{fig:reward_ablation}.

\subsection*{Safe multi-agent coordination at scale}

To evaluate scalability beyond training conditions, we conducted self-evaluation experiments where identical policies race against identical copies of themselves with progressively increasing competitor counts, from solo to 8-player races in Figure~\ref{fig:self_eval}A and \ref{fig:self_eval}B. This protocol tests generalization to agent densities exceeding the maximum of four opponents encountered during training. Single-agent policies exhibit severe performance degradation as the density of the competitors increases. Without opponent awareness, these policies select identical racing lines, resulting in a collision probability of 85\% at the first gate in four-player configurations. This failure mode highlights the fundamental inadequacy of opponent-agnostic approaches for multi-agent deployment.

Independent PPO agents, despite training with opponent observations, exhibit a different failure mode. Agents from the same training run achieve near-perfect collision avoidance through implicit coordination, converging to complementary racing lines visualized in Figure~\ref{fig:self_eval}C. However, this coordination does not generalize: when each independently learned policy is evaluated against itself, race completion drops to approximately 60\%. This brittleness indicates overfitting to specific co-training partners rather than learning robust interaction strategies.

Both fictitious self-play and league-play maintain approximately 90\% race completion at four agents, the configuration used during training, as both methods train against historical checkpoints of their own policies. However, their generalization diverges at higher agent densities: league-play sustains higher completion rates as competitor count increases, gradually decreasing to 56\% at eight competing agents. This advantage comes from exposure to various types of opponents during training, including single-agent and independent PPO policies, which produces strategies robust to varied interaction patterns. Crash rate analysis shown in Figure~\ref{fig:self_eval}B confirms that league-play achieves the lowest collision rates across all categories, demonstrating that opponent diversity during training translates directly into safer multi-agent coordination. Averaged across all agent counts, league-play reduces the average crash rate by over 50\% relative to single-agent policies (23.30\% versus 77.99\%).

\subsection*{Emergent anticipatory behaviors}

Understanding what policies learn is essential for deploying autonomous systems safely alongside humans. To interpret our learned policies, we analyze the critic network used during multi-agent training. By systematically varying components of the state input, such as ego position or opponent positions, we can visualize how the learned value function responds to different competitive scenarios. This approach reveals the strategic reasoning that emerges from interaction-based training and provides insight into how agents assess risk in the presence of competitors.

\begin{figure*}[htbp]
    \centering
    \includegraphics[width=\textwidth]{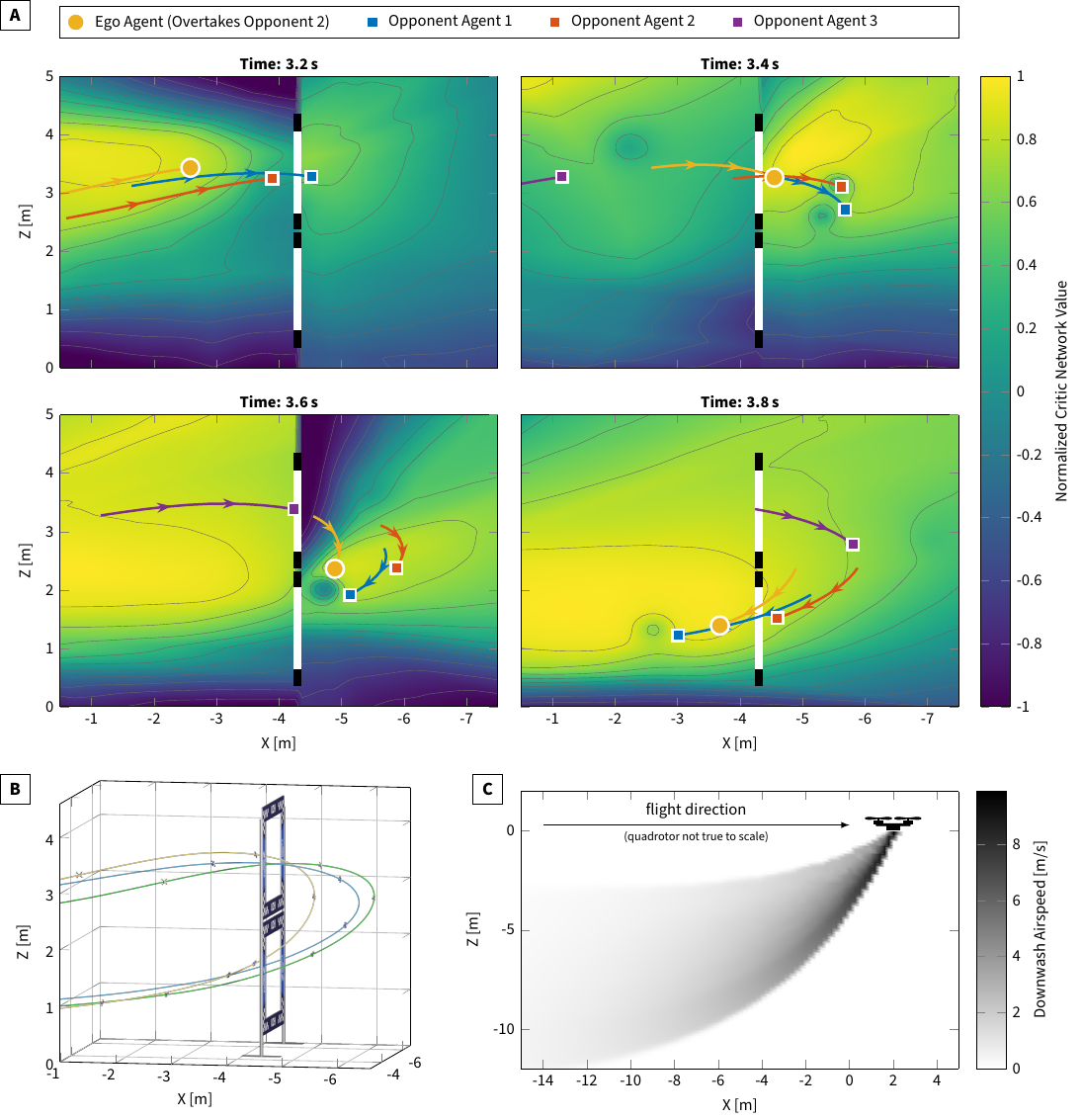}
    \caption{\textbf{A}, Learned value function visualized by varying the ego agent position in the $(X,Z)$ plane while fixing opponent positions. Trajectories are from a real-world overtaking maneuver at the Split-S gate, where competitors pass through an upper gate followed by a lower gate. The ego agent (orange) overtakes opponent 2 (red) across the four time steps shown. 
    Low-value regions near opponents reflect learned collision avoidance, while high values indicate regions of greater expected reward. For positions beyond the next gate, the visualization assumes the gate has been passed to avoid artificially low values from the agent seeking to return. \textbf{B}, Corresponding real-world trajectories showing the ego agent overtaking opponent 2 while avoiding opponent 1. \textbf{C}, Particle-based downwash simulation for a quadrotor moving in the positive $X$ direction. This aerodynamic interaction model is integrated into the training simulator, allowing agents to experience the physical effects of nearby competitors.}
    \label{fig:overtake_downwash}
\end{figure*}

Figure~\ref{fig:overtake_downwash}A shows the value function at four time steps during a real-world overtaking maneuver, obtained by sweeping the ego agent's position in the $(X,Z)$ plane while opponent world positions and all other kinematic states remain fixed. The resulting field encodes the expected return for occupying each $(X,Z)$ position under the current configuration, rather than a destination preference. The value function exhibits clear spatial structure reflecting learned awareness of collision risk: regions near opponents show reduced values, while regions clear of opponents and aligned with race progress retain high values. Importantly, this awareness is anticipatory rather than purely reactive. Low-value regions extend ahead of opponent positions along their projected trajectories, indicating that agents learn to reason about future collisions rather than responding only to immediate proximity, which can be interpreted as implicitly learning a world-model of the interactions.

The corresponding real-world trajectories in Figure~\ref{fig:overtake_downwash}B demonstrate these behaviors in practice. At the Split-S gate, requiring passage through an upper gate followed by a lower gate, the ego agent (orange) executes a successful overtake of opponent 2 (red) while maintaining safe separation from opponent 1. The value function snapshots reveal how the agent's assessment evolves throughout this maneuver: as the ego agent commits to the overtake, low-value regions track opponent positions, guiding the policy toward collision-free trajectories without sacrificing competitive positioning. We observed several successful overtakes across the real-world races, although such events remain infrequent given that competing policies fly close to time-optimal trajectories.

Safe coordination at high speeds also requires accounting for aerodynamic interactions between vehicles. We integrate an approximate downwash simulation into the training environment shown in Figure~\ref{fig:overtake_downwash}C, exposing agents to thrust disturbances from nearby competitors. This physical coupling encourages learned policies to maintain safe separation distances that account for aerodynamic effects beyond collision geometry alone. Details of the downwash model are provided in Materials and Methods.

\section*{Discussion}

This work demonstrates that multi-agent reinforcement learning, grounded in diverse self-play and physical interaction modeling, can achieve safe high-speed coordination in shared environments. By training against a league of opponents with varied strategies, our policies develop robust collision avoidance that generalizes beyond training conditions, enabling competitive performance against champion-level human pilots while maintaining consistent safety margins. These results establish autonomous quadrotor racing as a proving ground for multi-agent systems that must operate alongside humans.
A central finding is that interaction-aware training fundamentally changes how policies behave under competitive pressure. Human pilots, when trailing competitors, adopt increasingly aggressive strategies that elevate collision risk. Our learned policies, by contrast, maintain consistent safety margins, whether leading or trailing. This behavioral difference has important implications beyond racing: in any shared environment where humans and autonomous systems interact, predictable and conservative behavior from autonomous agents is essential for safe integration. The ability to train policies that internalize such behavior through self-play, without explicit safety constraints, suggests a promising path toward autonomous systems that are safe by design rather than by restriction.

The generalization properties of league-play trained policies merit particular attention. Training against diverse opponents, including single-agent policies that ignore competitors and independent learners that overfit to specific partners, produces robust strategies to novel interaction patterns. This diversity is essential: fictitious self-play alone exhibits degraded performance at higher agent densities, likely due to limited exposure to high-risk strategies employed by single-agent policies that prioritize speed over safety. Independently learned policies further expand the behavioral distribution, as joint optimization produces agents with distinct racing lines and interaction patterns that differ substantially from self-play dynamics. League-play maintains competitive performance even with eight simultaneous competitors, twice the maximum encountered during training. For real-world deployment, where the space of possible interactions is vast and unpredictable, such generalization is not merely desirable but necessary. Although the deployed policies share a single network architecture, both the training league and the large-scale evaluation place the framework in direct contact with heterogeneous policy populations spanning multiple learning paradigms, providing empirical evidence that interaction-aware behavior emerges from behaviorally distinct agents rather than only among homogeneous copies.

The Perceiver-based attention encoder proves essential for translating opponent observations into safe behavior. Our ablation reveals that naively incorporating opponent information without appropriate architectural processing degrades performance below single-agent baselines. This finding underscores a broader point: multi-agent learning requires not only appropriate training paradigms but also neural architectures capable of representing variable numbers of dynamic entities. The permutation-invariant structure of attention mechanisms provides a natural solution, enabling policies that scale gracefully as the number of competitors changes.

While these results establish the viability of multi-agent reinforcement learning for safe high-speed coordination, several directions for future work emerge.
Our policies currently act on full state estimates from motion capture, with position and velocity measurements of every agent, though we do not idealize this state: during training, we inject measurement noise and apply opponent-observation dropout. Moving to vision-based perception, as demonstrated for single-agent racing~\cite{kaufmann2023champion}, would remove this dependence and enable deployment outside instrumented spaces, while bringing autonomous and human pilots under comparable sensing constraints~\cite{geles2024demonstrating}. These constraints raise additional challenges, as a limited field of view leaves opponents only partially observable.

Looking ahead, the framework presented here extends naturally to domains beyond racing. Warehouse logistics, urban air mobility, and search and rescue all require multiple autonomous agents operating safely in shared physical space, often alongside humans. The core insight that training in diverse multi-agent settings produces generalizable and safe interaction strategies should transfer to these domains. League-based training could incorporate not only autonomous opponents but also models of human behavior, further improving the ability of learned policies to anticipate and accommodate human actions.
We recognize that capabilities enabling tight formation flight, strategic maneuvering, and robustness to physical disturbances could have applications beyond their intended scope. This research was conducted for civilian and sporting applications, and for the scientific advancement of embodied AI.
We encourage responsible adaptation of these methods and discourage their use in applications that could in any way cause harm to humans. Beyond these considerations, learned policies that reason about future collisions through training rather than explicit rules present challenges for accountability. In human-in-the-loop deployment, establishing responsibility for unintended harm requires understanding how policies assess risk. The value function analysis presented in this work offers one path forward: by visualizing how agents evaluate proximity to opponents and anticipate collision regions, we can inspect the learned representations underlying safety-critical decisions. Developing such interpretability tools will be essential for responsible deployment of interaction-aware autonomous systems in multi-agent environments.

More broadly, this work contributes to a growing body of evidence that competitive self-play, long successful in game-playing AI~\cite{tesauro1995temporal, silver2016mastering, silver2017mastering, vinyals2019grandmaster, berner2019dota}, can produce meaningful advances in physical robotics. The key enablers are simulation environments rich enough to capture relevant physical interactions and training frameworks diverse enough to prevent strategic overfitting. As simulation fidelity continues to improve and compute costs decrease, the scope of behaviors achievable through self-play will expand accordingly. We anticipate that multi-agent learning will become an essential tool for developing autonomous systems capable of safe, effective operation in the complex and dynamic environments that define real-world deployment.

\section*{Materials and Methods}

\subsection*{Simulation environment}

We train and evaluate policies in simulation using Flightmare~\cite{song2020flightmare} integrated with the Agilicious framework~\cite{foehn2022agilicious}, which provides high-fidelity dynamics for large-scale reinforcement learning. We extended the simulator to support multiple quadrotors simultaneously, enabling collision detection and aerodynamic interactions between agents. Multi-agent training is parallelized across 144 environments, each simulating four quadrotors, to accelerate data collection while exposing agents to varied multi-agent interaction dynamics.
The individual quadrotor dynamics are simulated as
\begin{equation}
\label{eq:3d_quad_dynamics}
\dot{\bm{x}} =
\begin{bmatrix}
\dot{\bm{p}}_{\wfr\bfr} \\  
\dot{\bm{q}}_{\wfr\bfr} \\
\dot{\bm{v}}_{\wfr} \\
\dot{\boldsymbol\omega}_\bfr \\
\dot{\boldsymbol\Omega}
\end{bmatrix} = 
\begin{bmatrix}
\bm{v}_\wfr \\  
\bm{q}_{\wfr\bfr} \cdot \begin{bmatrix}
0 \\ \bm{\omega}_\bfr/2\end{bmatrix} \\
\frac{1}{m} \Big(\bm{q}_{\wfr\bfr} \odot (\bm{f}_\text{prop} + \bm{f}_\text{aero})\Big)+\bm{g}_\wfr  \\
\bm{J}^{-1}\big( \boldsymbol{\tau}_\text{prop} + \boldsymbol{\tau}_\text{aero}  - \boldsymbol{\omega}_\bfr \times \bm J \boldsymbol{\omega}_\bfr\big) \\
\frac{1}{k_\text{mot}} \big(\boldsymbol\Omega_\text{ss} - \boldsymbol\Omega \big)
\end{bmatrix} \; ,
\end{equation}
where $\odot$ represents quaternion rotation, $\bm{p}_{\wfr\bfr}$, $\bm{q}_{\wfr\bfr}$, $\bm{v}_{\wfr}$, and $\boldsymbol\omega_\bfr$ denote the position, orientation quaternion, inertial velocity, and body rates of the quadrotor, respectively. The motor time constant is $k_\text{mot}$, and the motor speeds $\Omega$ and $\Omega_\text{ss}$ are the actual and steady-state motor speeds, respectively. The matrix $\bm J$ is the inertia, and $\bm{g}_\wfr$ denotes the gravity vector. 
The force and torque contributions of the propeller and motor unit, as well as aerodynamic effects are denoted by $\bm f_\text{prop}, \boldsymbol{\tau}_\text{prop}$ and $\bm f_\text{aero}, \boldsymbol{\tau}_\text{aero}$. Propeller forces and torques $(\bm{f}_\text{prop}, \boldsymbol{\tau}_\text{prop})$ are computed using a quadratic thrust model with data-driven augmentation for computational efficiency during training~\cite{bauersfeld2021neurobem, kaufmann2023champion}. Aerodynamic forces and torques $(\bm{f}_\text{aero}, \boldsymbol{\tau}_\text{aero})$ capture drag effects at high speeds.
For multi-agent training, the simulator runs multiple quadrotor instances in parallel, with each agent subject to identical dynamics. Aerodynamic interactions between agents are modeled through a particle-based downwash simulation described in the following section.

\subsection*{Downwash simulation and interactions}

Multi-player racing requires modeling aerodynamic interactions between closely flying quadrotors, as propeller downwash creates significant disturbances that affect flight dynamics during close-proximity maneuvers. We developed a particle-based simulation that provides a computationally tractable approximation of these effects.
Building on characterizations of quadrotor airflow as turbulent jets~\cite{bauersfeld2025robotics}, our approach models downwash through particles emitted beneath each quadrotor proportional to thrust output. The initial particle velocity $v_i$ follows momentum theory for propeller-induced flow:
\begin{equation}
v_i = \sqrt{\frac{T}{2 \rho A_{\mathrm{prop}}}}
\end{equation}
where $T$ is the instantaneous thrust, $\rho$ is air density, and $A_{\mathrm{prop}}$ is the propeller disk area. In every simulation step, 96 particles are emitted in a conical distribution beneath each rotor, with velocities that spread spatially and decay temporally according to turbulent jet characteristics. The local airspeed due to downwash can then be obtained by interpolating from nearby particles. The resulting wind field (shown in Figure~\ref{fig:overtake_downwash}C for a drone flying at a constant velocity) creates local wind disturbances that modify the aerodynamic forces experienced by nearby agents.
These wind fields are integrated into each quadrotor's drag model by computing the relative airspeed between the vehicle and local flow. Agents flying through opponent downwash experience altered aerodynamic forces, requiring compensatory control responses. While this approach simplifies the full complexity of the aerodynamic interactions and the airflow model is only a very coarse approximation, it captures the dominant effects relevant to close-proximity flight. We further verify the importance of modeling these aerial interactions for training multi-agent policies in an experiment reported in Supplementary Fig.~\ref{fig:downwash_toyexample}.

\subsection*{Multi-agent RL framework}

\subsubsection*{Problem formulation}

We formulate multi-agent racing as a Markov game~\cite{littman1994markovgame}, the multi-agent extension of the Markov decision process (MDP) used in single-agent reinforcement learning. The environment is shared between $N$ agents, where at every time-step $t$ each agent $i$ observes its own state $s_{t, \text{ego}}$ and the states of other agents $s_{t, \text{other}} = \{s_{t, \text{other}}^{(j)}\}_{j \neq i}$. Each agent seeks to maximize its expected cumulative reward given its own policy $\pi_{\text{ego}}$ and the policies of other agents $\pi_{\text{other}}$:
\begin{equation}
J(\pi_{\text{ego}}) = \mathbb{E}_{\pi_{\text{ego}}, \pi_{\text{other}}} \left[ \sum_{t=0}^{T} \gamma^t r_t \mid s_{t, \text{ego}}, s_{t, \text{other}} \right]
\end{equation}
where $\gamma$ is the discount factor and $r_t$ is the reward at time-step $t$. The expectation is taken over trajectories induced by both the ego policy and opponent policies. Unlike single-agent settings, the optimal policy depends on the behavior of other agents, motivating training against diverse opponents.

\subsubsection*{Observation and action spaces}

The ego state $s_{t, \text{ego}}$ follows the formulation of~\cite{song2023reaching}:
\begin{equation}
s_{t, \text{ego}} = \left[ \mathbf{p}, \mathbf{v}, \mathbf{R}, \mathbf{g}_{\text{corners}}, \mathbf{g}_{\text{next}} \right]
\end{equation}
where $\mathbf{p} \in \mathbb{R}^3$ is the ego position, $\mathbf{v} \in \mathbb{R}^3$ is the ego velocity, $\mathbf{R} \in \mathbb{R}^{3 \times 3}$ is the ego rotation matrix, $\mathbf{g}_{\text{corners}} \in \mathbb{R}^{12}$ contains distances from the ego agent to the four corners of the next gate, and $\mathbf{g}_{\text{next}} \in \mathbb{R}^{12}$ encodes the relative difference in gate corner distances between the next two consecutive gates. Opponent observations for each competitor $j$ are defined as:
\begin{equation}
s_{t, \text{other}}^{(j)} = \left[ \mathbf{p}_{\text{rel}}^{(j)}, \mathbf{v}_{\text{rel}}^{(j)} \right]
\end{equation}
where $\mathbf{p}_{\text{rel}}^{(j)} \in \mathbb{R}^3$ and $\mathbf{v}_{\text{rel}}^{(j)} \in \mathbb{R}^3$ denote the relative position and velocity of opponent $j$ with respect to the ego agent.

The action space consists of collective thrust and body rate commands $\mathbf{a}_t = \left[ c, \omega_x, \omega_y, \omega_z \right] \in \mathbb{R}^4$,
where $c$ is the mass-normalized collective thrust and $\boldsymbol{\omega} = [\omega_x, \omega_y, \omega_z]$ are the commanded body rates. This control interface provides direct access to the vehicle's agility while relying on a low-level controller to track the commanded rates~\cite{kaufmann2023champion, song2023reaching}. The commanded collective thrust is bounded by the maximum achievable motor thrust (14~N total), ensuring policy outputs remain within physically realizable limits; see Supplementary Materials for hardware specification details.

\subsubsection*{Policy architecture}

To handle variable numbers of competitors, opponent observations $\{s_{t, \text{other}}^{(j)}\}_{j=1}^{N-1}$ are processed through a Perceiver-based attention encoder~\cite{jaegle2021perceiver}. The encoder uses $4$ learned latent queries, each attending to the full set of opponent observations via $4$-head cross-attention with a head dimension of $32$, producing a fixed-dimensional, permutation-invariant representation regardless of the number of competitors. To isolate the contribution of this architectural choice, we evaluate an ablation that removes the Perceiver encoder and instead concatenates a fixed-length opponent representation directly with the ego state. This baseline accommodates up to 10 opponents by zero-padding absent observations, yielding a flattened input vector that lacks permutation invariance and scales poorly with competitor count. The encoding is concatenated with the ego state and passed to recurrent actor and critic networks. 

Both actor and critic networks comprise a single LSTM layer with $256$ hidden units followed by two fully connected layers with $512$ units each. LeakyReLU activations are used throughout the network, with the exception of the actor output layer, which employs a hyperbolic tangent to bound actions within a normalized range. The recurrent architecture enables agents to maintain beliefs about opponent intentions and race dynamics across time steps, capturing temporal dependencies that are critical for anticipatory behavior. 
An architectural overview is provided in the Supplementary Fig.~\ref{fig:method}.

\subsubsection*{League-play training}
Policies are trained using a recurrent variant of Proximal Policy Optimization (PPO)\cite{schulman2017proximal} for 5,500 iterations totaling 200 million environment interactions, requiring approximately 27 hours of wall-clock time on a single NVIDIA RTX 4090 GPU; full hyperparameters and additional training details are reported in the Supplementary Materials. Our implementation builds on Stable-Baselines3~\cite{raffin2021stable-baselines3}, extended to support multi-agent training with league-based self-play and independent learning configurations.
Training follows a league-play paradigm inspired by~\cite{vinyals2019grandmaster}. In each training iteration, a single learning agent is paired with opponents sampled from two distinct pools: historical checkpoints from its own training (fictitious self-play) and a fixed league of diverse opponents. With probability 75\%, opponents are drawn from the agent's own checkpoint history; otherwise, opponents are sampled from the league pool.
For fictitious self-play, checkpoints are saved every 100 iterations, accumulating approximately 55 snapshots over the course of training. To ensure the learning agent faces increasingly competent opponents as training progresses, checkpoint selection follows a power-law distribution $P(k) = \frac{k^{\alpha}}{\sum_{j=1}^{K} j^{\alpha}}, \quad k \in \{1, \ldots, K\}$ where $k$ is the checkpoint index ordered chronologically such that $k=1$ corresponds to the earliest checkpoint and $k=K$ to the most recent, $K$ is the total number of saved checkpoints, and $\alpha = 0.9$ is the power-law exponent. This weighting biases sampling toward more recent checkpoints, which exhibit stronger performance, while maintaining non-zero probability of encountering earlier strategies to prevent catastrophic forgetting of responses to suboptimal play.
The league pool contains 20 fixed policies capturing diverse competitive behaviors: four single-agent policies optimized purely for lap time without opponent awareness, and sixteen independently trained PPO policies drawn from four jointly trained agents at various stages of training. Single-agent opponents contribute aggressive, high-risk racing strategies, while independent PPO policies introduce varied racing lines and interaction patterns emerging from simultaneous multi-agent optimization. In the independent PPO baseline, four agents are trained simultaneously within each environment, each maintaining separate policy and value networks; rollouts are collected from shared races in which all four agents act concurrently, and each network is updated using PPO from its own observed transitions.

This diversity exposes the learning agent to competitive scenarios absent from self-play alone, where opponents share similar strategic tendencies. While joint optimization in independent PPO produces coordinated behavior among co-trained agents, such coordination fails to generalize beyond specific training partners, as demonstrated in our experimental evaluation.

\subsubsection*{Curriculum learning}

Training multi-agent policies from scratch presents significant exploration challenges, as agents must simultaneously learn aggressive flight control and opponent-aware behavior. We address this through a staged curriculum that progressively increases the difficulty of the task.

Initially, opponent agents are invisible and non-collidable, allowing the ego agent to first acquire proficient single-agent racing behavior. As training progresses, opponents become visible, and physical interactions are enabled, shifting the learning objective toward multi-agent coordination. The dimensions of the gates are initialized at the nominal size of $2\times$ and gradually reduced to $1\times$, relaxing the precision requirements during early exploration.

To ensure robust performance from race start conditions, we employ curriculum-based initialization using an initial state buffer. During training, episodes are initially initialized predominantly from states sampled on the track to encourage exploration. As training progresses, the probability of initialization at the start of the race increases from $5\%$ to $95\%$, ensuring that the final policy performs reliably under competition conditions where all agents begin simultaneously.

\subsubsection*{Domain randomization}

To bridge the simulation-to-reality gap and prevent policies from memorizing control sequences, we employ domain randomization across vehicle dynamics and initial conditions~\cite{song2023reaching, kaufmann2023champion}. Despite access to accurate dynamics models, drone-to-drone variation and complex aerodynamic effects necessitate robust policies that generalize across parameter ranges.

During training, we randomize thrust and torque coefficients by $\pm 10\%$, body drag by $\pm 10\%$, inertia by $\pm 10\%$, and mass by $\pm 5\%$. A random actuation delay of up to $\unit[40]{ms}$ is applied to account for communication and processing latencies. The initial conditions are sampled uniformly around the nominal starting configuration: the position varies by $\unit[\pm 0.5]{m}$ in the horizontal plane and $\unit[\pm 0.3]{m}$ vertically, the velocity by $\unit[\pm 0.5]{m/s}$ per axis, the attitude by $\unit[\pm 20]{deg}$, and the body rates by $\unit[\pm 25]{deg/s}$. This randomization ensures that the learned policies respond to environmental feedback rather than executing open-loop trajectories, enabling successful zero-shot transfer to physical hardware.

\subsubsection*{Reward function}

Following established practice in learned quadrotor control~\cite{song2023reaching}, we employ a dense, shaped reward function that provides continuous feedback throughout each episode. The reward function extends previous work with additional terms to account for multi-agent interaction and competitive racing. For non-terminal time steps, the per-step reward is defined as
\begin{equation}
r_t = r_t^{\text{prog}} - r_t^{\omega} - r_t^{\text{prox}} + r_t^{\text{rank}} .
\end{equation}
The individual reward components are given by
\begin{align}
r_t^{\text{prog}} &= \lambda_1 ( d_{t-1} - d_t ), \\[4pt]
r_t^{\omega} &= \lambda_2 \| \boldsymbol{\omega}_t \|, \\[4pt]
r_t^{\text{rank}} &= \lambda_3 \frac{N - (\text{rank}_t - 1)}{N}, \\[4pt]
r_t^{\text{prox}} &=
\begin{cases}
(\lambda_4 \| \mathbf{v}_t \| + 1)\exp(- \lambda_5 \tilde{d}_t), & \text{if } d_{\text{opp}} < 2 d_{\text{col}}, \\
0, & \text{otherwise}.
\end{cases}
\end{align}
Here, $d_t$ denotes the distance to the next gate center, $\boldsymbol{\omega}_t$ is the body rate, $\mathbf{v}_t$ is the velocity, $N$ is the number of agents, and $\text{rank}_t \in \{1,\ldots,N\}$ is the current ranking based on race progress, derived from the number of passed gates and distance to the next gate. The proximity term activates when the distance to the nearest opponent $d_{\text{opp}}$ falls below twice the collision radius $d_{\text{col}} = \unit[0.1]{m}$, with normalized separation $\tilde{d}_t = (d_{\text{opp}} - d_{\text{col}})/d_{\text{col}}$. 
This collision radius is chosen to match the physical extent of the quadrotor platform.

Collisions with walls, gates, or other agents terminate the episode. Rather than including collision penalties in the per-step reward, a terminal penalty is applied at the final time step: wall and inter-agent collisions incur a penalty proportional to impact velocity $\|\mathbf{v}_t\|$, with the inter-agent term additionally scaled by $\lambda_4$ to maintain consistency with the velocity-dependent proximity penalty, while gate collisions incur a penalty proportional to the squared traversal error  $e_{\text{gate}}^2$. Both cases include a constant offset of $-1$ to discourage any collision regardless of severity. To account for scenarios where vehicles make brief contact but remain airworthy, 10\% of inter-agent collisions are randomly selected to be non-terminal, allowing the episode to continue after the collision penalty is applied. This stochastic termination reflects the reality that not all mid-air contacts result in catastrophic failure, encouraging policies to recover from minor collisions rather than treating any contact as terminal. When an inter-agent collision terminates the episode for both involved agents, final rankings are determined by their relative race progress immediately prior to impact, computed from the number of gates passed and proximity to the next gate.

The progress term $r_t^{\text{prog}}$ incentivizes advancement toward the next gate, while the body rate penalty $r_t^{\omega}$ promotes smooth and dynamically feasible flight. The proximity penalty $r_t^{\text{prox}}$ provides a continuous shaping signal that discourages unsafe close approaches to opponents before physical contact occurs, and the ranking reward $r_t^{\text{rank}}$ encourages competitive behavior by rewarding relative race position. This reward structure yields policies that balance aggressive racing with safety-conscious decision making. The coefficient values are provided in the Supplementary Material, along with an ablation study on the proximity coefficient $\lambda_4$ (Supplementary Fig.~\ref{fig:reward_ablation}) that characterizes the resulting trade-off between lap time and race completion.

\subsection*{Real-world deployment}
Real-world experiments were conducted in an indoor flight arena equipped with a 42-camera motion capture system providing state estimation at up to $\unit[400]{Hz}$. The racetrack comprises seven gates arranged in an approximately 75-meter circuit, including a Split-S maneuver requiring sequential passage through vertically stacked gates. This configuration demands aggressive three-dimensional flight with rapid altitude changes, representative of professional drone racing courses.

All quadrotors are identical racing platforms based on the Agilicious framework~\cite{foehn2022agilicious}, with a mass of $\unit[220 \pm 3]{g}$ and a thrust-to-weight ratio of 6.5 and 3-inch propeller diameter. This standardization ensures that performance differences arise from policy behavior rather than hardware variation. 
The policy outputs collective thrust and body rate commands, which are tracked by a low-level controller running onboard each vehicle. Although each agent computes commands based only on its own observations, all policy inference runs offboard on a single desktop computer, highlighting the lightweight computational requirements of the architecture.
Autonomous agents execute learned policies at $\unit[50]{Hz}$ control frequency, receiving ego state and opponent observations from the motion capture system. Despite the simulation dynamics model incorporating real-world data, no further policy fine-tuning was performed; all real-world flights use zero-shot transfer.

Race starts were initiated via audio signal to ensure identical timing information for all participants. To match typical human reaction times, autonomous agents begin execution with a random delay sampled between 50 and $\unit[250]{ms}$ following the start signal. In mixed human-AI races, the human pilot selected their starting position subject to the constraint that distances to other agents and to the first gate remained equal across all competitors. Agents are positioned along an arc at equal distance to the first gate, with $\unit[1]{m}$ spacing between adjacent starting positions. In mixed human-AI races, the human pilot selected their position along this arc, maintaining equal distance to the gate and neighboring competitors. While autonomous agents initialize from a stable hover, the human pilot preferred launching from a stationary podium, eliminating the cognitive load of maintaining position during the audio start signal countdown.

The human pilot was familiar with the track layout due to several prior experiments~\cite{kaufmann2023champion, song2023reaching} and was provided two hours of practice flights before recorded trials to ensure performance representative of expert-level racing. This protocol ensures that observed differences between human and autonomous performance reflect strategic and perceptual factors rather than track unfamiliarity.

Time-trial performance confirms the human pilot's champion-level skill: measured from first gate passage within longer multi-lap sessions where approach speed was already established, his best single lap ($\unit[5.19]{s}$) and three-lap ($\unit[15.69]{s}$) times closely match our policy's best recorded times ($\unit[5.10]{s}$ and $\unit[15.62]{s}$). This near-parity in isolated flight establishes a matched baseline, ensuring that performance differences observed in multi-agent races reflect competitive dynamics rather than disparities in fundamental piloting capability.

\subsection*{Human pilot impressions}

The following statements summarize the qualitative impressions of expert pilot Marvin Schaepper after racing against autonomous agents. He highlighted the agents' ability to maintain extremely tight formations, noting that such close-proximity flight would be difficult for human pilots to sustain. In addition, he reported that densely packed groups increased cognitive workload, making it challenging to anticipate and execute overtaking maneuvers when several opponents were flying in close proximity. With respect to race dynamics, he emphasized the importance of the start phase, observing that falling behind early often forces human pilots to adopt riskier trajectories in an attempt to recover position, which in turn increases the likelihood of crashes. These impressions are consistent with our quantitative findings and support the conclusion that human pilots adapt their risk tolerance based on race context, while the learned policies exhibit more stable and safe behavior across varying conditions.

\section*{Acknowledgments}
We thank Ben Moran, Simon Osindero, Amanda Prorok, Petros Koumoutsakos, Yunfan Ren, Rudolf Reiter, Jiaxu Xing, Elie Aljalbout, Angel Romero, Yifan Zhai, Dimosthenis Angelis, and Yannick Armati for their feedback, insightful discussions, and assistance with the real-world experiments.
We thank Drew Hanover and Yunlong Song for their early exploratory work on this research direction during their time at the lab. 
We are grateful to Marvin Schaepper for his expert participation in the human-AI racing trials and for sharing his qualitative impressions of competing against multiple autonomous agents.

\paragraph*{Funding:}
This work was supported by the European Union’s Horizon Europe Research and Innovation Programme under grant agreement No. 101120732 (AUTOASSESS), the European Research Council (ERC) under grant agreement No. 864042 (AGILEFLIGHT), and the UZH Candoc Grant, grant no. FK-25-010.
\paragraph*{Author contributions:}
I.G. formulated the main ideas, implemented the system, performed all experiments and data analysis, and wrote the paper. L.B. contributed to the main ideas, implementation, experiments, data analysis, multimedia materials. M.W. contributed to the main ideas, project conception, the experimental design, and paper writing. D.S. contributed to the main ideas, project conception, paper writing, and provided funding.

\paragraph*{Competing interests:}
There are no competing interests to declare.

\paragraph*{Data and materials availability:} 
All (other) data needed to evaluate the conclusions in the paper are present in the manuscript or the Supplementary Material. Correspondence and requests for materials should be addressed to I.G. Motion capture recordings of the racing trajectories are available at \url{https://doi.org/10.5281/zenodo.20190159}

\paragraph{Research Ethics:}
The study has been conducted in accordance with the Declaration of Helsinki. The study protocol is exempt from review by an ethics committee according to the rules and regulations of the University of Zurich, because no health-related data has been collected. The participants gave their written informed consent before participating in the study.

\paragraph*{Supplementary materials:}~\\
\noindent Supplementary Text\\
Figs. S1 to S3\\
Tables S1 to S4\\
\href{https://youtu.be/dumWP-cwE3w}{Movie S1}\\
\href{https://rpg.ifi.uzh.ch/marl}{Project Website}\\

\vspace{-0.5cm}

\bibliography{references} %

\clearpage
\newpage

\setcounter{table}{0}
\makeatletter 
\renewcommand{\thetable}{S\@arabic\c@table}
\makeatother

\setcounter{figure}{0}
\makeatletter 
\renewcommand{\thefigure}{S\@arabic\c@figure}
\makeatother

\setcounter{algorithm}{0}
\makeatletter 
\renewcommand{\thealgorithm}{S\@arabic\c@algorithm}
\makeatother

\setcounter{section}{0}
\makeatletter 
\renewcommand{\thesection}{S\@arabic\c@section}
\makeatother

\section*{Supplemetary Material}

\subsection*{Reward coefficients}

Table~\ref{tab:reward_coefficients} lists the coefficients used in the reward function defined in the Materials and Methods section. The progress coefficient $\lambda_1$ provides the primary learning signal, while the remaining coefficients are scaled to balance competing objectives without dominating the progress reward. The relatively small values for ranking ($\lambda_3$) and opponent proximity ($\lambda_4$) encourage competitive behavior and safe distancing without overriding the fundamental gate-progress objective.

\begin{figure}[htbp]
    \centering
    \includegraphics[width=0.4\textwidth]{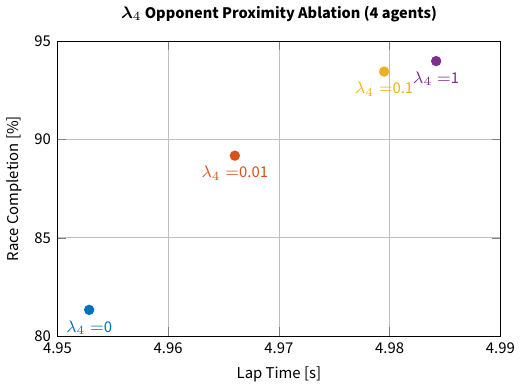}
    \caption{Ablation on proximity coefficient $\lambda_4$, showing the trade-off between lap time and race completion as fraction of gates passed over three laps. Four league-play policies per coefficient value are evaluated in four-agent self-evaluation. $\lambda_4 = 0.01$ balances speed and safety.}
    \label{fig:reward_ablation}
\end{figure}

The proximity coefficient $\lambda_4$ scales the continuous penalty for approaching opponents and is therefore critical for balancing competitive performance against safety. This coefficient additionally scales the velocity-dependent terminal penalty upon inter-agent collision. Figure~\ref{fig:reward_ablation} shows the trade-off between lap time and race completion (fraction of gates passed over three laps) as $\lambda_4$ varies. For each value, four League-Play checkpoints were evaluated in four-agent self-evaluation races. Lower $\lambda_4$ yields faster but riskier flight, while higher values improve safety at the cost of speed. The selected value $\lambda_4=0.01$ maintains competitive lap times while completing over 89\% of races.

\begin{table}[!ht]
\centering
\caption{Coefficients for the reward function defined in Materials and Methods.}
\begin{tabularx}{0.75\linewidth}{lX}
\textbf{Parameter} & \textbf{Value} \\ \midrule
$\lambda_1$ (progress) & $1.0$ \\
$\lambda_2$ (body rate penalty) & $0.005$ \\
$\lambda_3$ (ranking) & $0.01$ \\
$\lambda_4$ (opponent proximity) & $0.01$ \\
$\lambda_5$ (opponent proximity exponent) & $7.0$ \\
\bottomrule
\end{tabularx}
\label{tab:reward_coefficients}
\end{table}

\subsection*{Training details}
Table~\ref{tab:training_hyperparameters} summarizes the PPO hyperparameters and network architecture used for training league-play policies. The learning rate follows a linear decay schedule from $3 \times 10^{-4}$ to $5 \times 10^{-5}$ over the first half of training, then remains constant. With $144$ parallel environments, each containing four quadrotors in the multi-agent setting (one quadrotor in the single-agent setting), and $250$ steps per rollout, each iteration collects $36{,}000$ transitions.
Training proceeds for $5{,}500$ iterations, yielding approximately $200$ million environment steps. 
League-play training with downwash simulation was performed using an NVIDIA RTX 4090 GPU and an Intel Core i9-14900KF CPU, requiring approximately 27 hours to complete (assuming pre-trained league opponents), with potential for further optimization. Unless otherwise specified, the parameters in Table~\ref{tab:training_hyperparameters} are shared across all training strategies evaluated in this work.

\begin{table}[!ht]
\centering
\caption{PPO hyperparameters and network architecture for league-play training.}
\begin{tabularx}{0.9\linewidth}{lX}
\textbf{Parameter} & \textbf{Value} \\ \midrule
learning rate & $3 \times 10^{-4}$ linear decay to $5 \times 10^{-5}$ \\
discount factor $\gamma$ & $0.985$ \\
GAE-$\lambda$ & $0.95$ \\
PPO epochs & $10$ \\ 
clip range & $0.2$ \\
entropy coefficient & $0.001$ \\ 
parallel environments & $144$ \\
rollout steps & $250$ \\
batch size & $36{,}000$ \\
episode length & $\unit[30]{s}$ \\
control frequency & $\unit[50]{Hz}$ \\
activation function & LeakyReLU \\
policy network MLP & $[512, 512]$ \\
value network MLP & $[512, 512]$ \\
LSTM hidden dimension & $[256]$ \\
Perceiver latent queries & $4$ \\
Perceiver attention heads & $4$ \\
Perceiver head dimension & $32$ \\
\bottomrule
\end{tabularx}
\label{tab:training_hyperparameters}
\end{table}

\subsection*{Hardware specifications}

Table~\ref{tab:hardware} lists the specifications of the Kolibri quadrotor platform used in all real-world experiments. The high thrust-to-weight ratio enables the aggressive maneuvers required for competitive racing, while the compact form factor permits close-proximity flight between multiple agents.

\begin{table}[!ht]
\centering
\caption{Kolibri quadrotor platform specifications.}
\begin{tabularx}{0.75\linewidth}{l>$C<$}
\textbf{Parameter} & \textbf{Value} \\ \midrule
mass [kg]& 0.220 \\
thrust-to-weight ratio & 6.5 \\
maximum thrust [N]& 14.0 \\
motor-to-motor distance [cm]& 11.8 \\ 
propeller diameter [cm]& 7.37 \\ 
motor time constant [s]&  0.033 \\
inertia (diagonal terms) [gm$^2$] & [0.14, 0.17, 0.21] \\
\bottomrule
\end{tabularx}
\label{tab:hardware}
\end{table}

\subsection*{Racetrack configuration}

Table~\ref{tab:racetrack} specifies the gate positions and orientations for the Split-S racetrack: seven gates in a 75-meter circuit, with Gates 4 and 5 vertically stacked to form the Split-S maneuver. 

All gates measure {$\unit[1.5]{m}$ $\times$ $\unit[1.5]{m}$}.
Agents initialize along an arc equidistant from the first gate, with $\unit[1]{m}$ spacing between adjacent positions. The reference starting position is $(-5.0, 4.7, 0.61)~\unit[]{m}$. The starting slots are randomized during training. For experiments exceeding four agents, positions shift by half a spacing unit per additional agent to maintain a symmetric distribution to the first gate.

\begin{table}[htb]
\centering
\caption{Split-S racetrack gate configuration. Positions are specified in a world coordinate frame with $z$ pointing upward, and yaw angles indicate gate orientation about the vertical axis.}

\begin{tabularx}{0.95\linewidth}{X|>$r<$>$r<$>$r<$>$r<$}
\textbf{Gate} & \text{x [m]} & \text{y [m]} & \text{z [m]} & \text{yaw [deg]} \\ \midrule
Gate 1 & -0.60 & -0.86 & 3.68 & -20 \\
Gate 2 & 9.00 & 6.45 & 1.05 & 0 \\
Gate 3 & 8.85 & -3.80 & 1.05 & -130 \\
Gate 4 (Split-S Top) & -4.30 & -5.60 & 3.40 & 180 \\
Gate 5 & -4.30 & -5.60 & 1.42 & 0 \\
Gate 6 & 4.50 & -0.45 & 1.05 & 80 \\
Gate 7 & -1.95 & 6.81 & 1.05 & -150 \\
\bottomrule
\end{tabularx}
\label{tab:racetrack}
\end{table}

\begin{figure*}[p]
\subsection*{Architecture overview} \ \\ \ \\
\centering
    \includegraphics[]{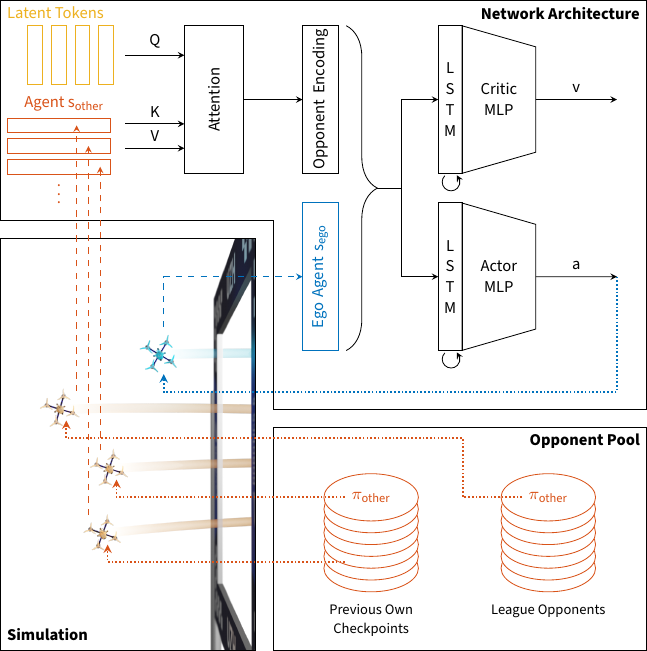}
    \caption{Multi-agent training framework. Each agent observes its own state $s_{\text{ego}}$ and the states of other agents $s_{\text{other}}$. Opponent observations serve as keys and values within a Perceiver-based attention encoder, which queries them with learned latent arrays to produce a compact representation regardless of the number of competitors. The encoded opponent representation is concatenated with the ego state and passed to recurrent actor and critic networks comprising LSTM and MLP layers. During training, opponent policies are sampled from an opponent pool that includes historical checkpoints of the agent's own policy alongside the league opponents, which comprise single-agent and independent PPO policies, providing diverse competitive behaviors.}
    \label{fig:method}
\end{figure*}

\enlargethispage{-10cm}
\pagebreak

\subsection*{Aerodynamic interactions in training}

To isolate the contribution of aerodynamic interaction modeling, we designed a controlled experiment with two quadrotors flying concentric circular trajectories of $\unit[3]{m}$ radius, vertically separated by $\unit[0.5]{m}$. Both agents execute identical policies trained with or without particle-based downwash simulation. The upper agent does not observe the lower agent, maintaining consistent flight regardless of proximity. The lower agent observes the upper agent, enabling response to anticipated aerodynamic disturbance if trained with downwash. The reward function was modified to encourage a constant velocity of $\unit[2.5]{m/s}$ by penalizing deviations from this reference. We evaluated three configurations: the lower agent flying alone, and following the upper agent with initial delays of $\unit[0.1]{s}$ and $\unit[0.5]{s}$, with $15$ flights per deployed configuration.

\begin{figure*}[!b]
    \centering
    \includegraphics[width=0.8\textwidth]{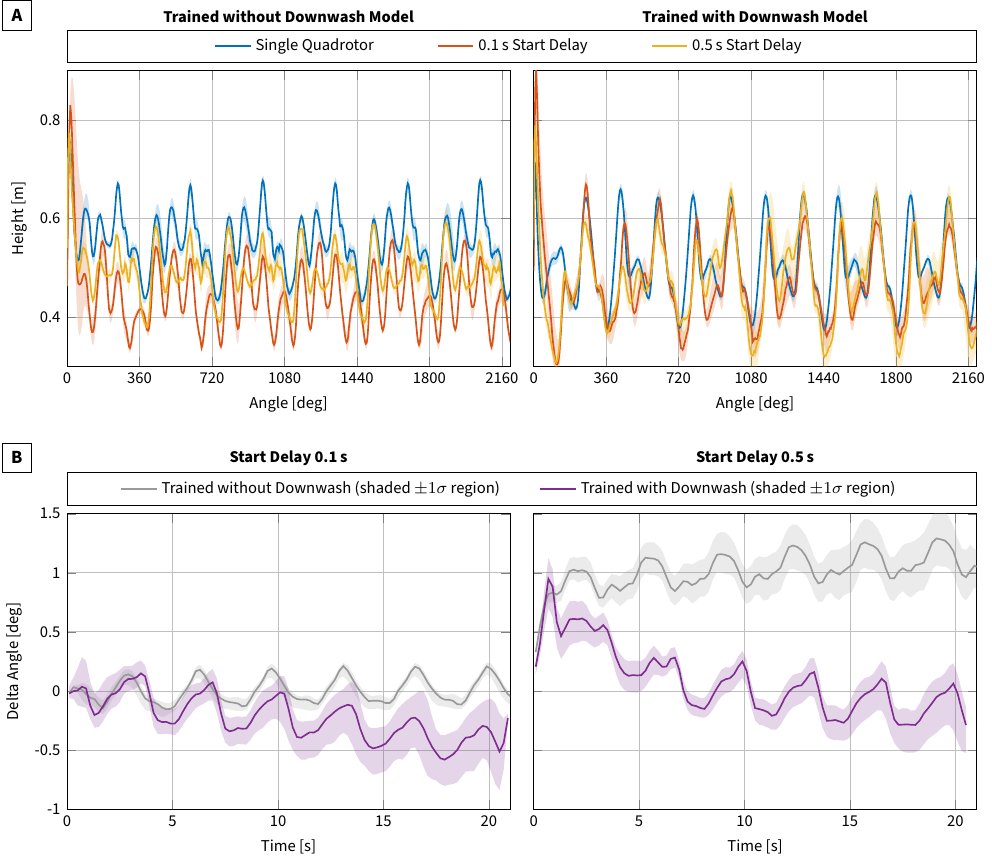}
    \caption{Effect of downwash modeling on learned behavior. Two quadrotors fly concentric circles ($\unit[3]{m}$ radius, $\unit[0.5]{m}$ vertical separation). \textbf{(A)} Altitude profiles for policies trained without (left) and with (right) downwash across solo flight, $\unit[0.1]{s}$ delay, and $\unit[0.5]{s}$ delay conditions. \textbf{(B)} Angular difference between upper and lower agent over time for $\unit[0.1]{s}$ (left) and $\unit[0.5]{s}$ (right) initial delay, comparing policies trained with and without downwash. Negative values indicate the lower agent leads. Interaction-aware policies maintain consistent altitude and converge to leading positions, while policies without downwash exhibit condition-dependent drift. Shaded regions denote variance across $15$ flights.}
    \label{fig:downwash_toyexample}
\end{figure*}

Figure~\ref{fig:downwash_toyexample}A reveals distinct behavioral strategies depending on training conditions. Policies trained without downwash (left) show altitude profiles that vary substantially across conditions: solo flight produces stable tracking, while following the upper agent results in degraded performance as the policy lacks incentive to avoid the wake region. By contrast, policies trained with downwash (right) maintain consistent altitude across all conditions, actively accelerating to avoid the turbulent wake.
The angular separation over time in Figure~\ref{fig:downwash_toyexample}B quantifies this behavioral difference. At $\unit[0.1]{s}$ initial delay (left), policies without downwash awareness remain near or behind the upper agent, while downwash-aware policies rapidly converge to negative angles, indicating a leading position. This pattern is more pronounced at $\unit[0.5]{s}$ delay (right): without downwash training, the agent drifts progressively behind, whereas downwash-aware policies consistently overtake the upper agent despite the larger initial gap. These results validate that the downwash simulation provides sufficient training signal for agents to learn physically meaningful avoidance behaviors.

\end{document}